\documentclass{article}

\usepackage{PRIMEarxiv}

\usepackage[utf8]{inputenc} 
\usepackage[T1]{fontenc}    
\usepackage{hyperref}       
\usepackage{url}            
\usepackage{booktabs}       
\usepackage{amsfonts}       
\usepackage{nicefrac}       
\usepackage{microtype}      
\usepackage{lipsum}
\usepackage{fancyhdr}       
\usepackage{graphicx}       
\graphicspath{{media/}}     

\usepackage{amssymb}
\usepackage{booktabs}
\usepackage{amsmath,amssymb,amsfonts}
\usepackage{mathtools}
\usepackage{algorithmicx}
\usepackage{algorithm}
\usepackage{algpseudocode}

\usepackage{graphicx}
\usepackage{textcomp}
\usepackage{graphics} 
\usepackage{epsfig}
\usepackage{algorithm}
\usepackage{mathabx}
\usepackage{enumerate}
\usepackage{enumitem} 
\usepackage[]{algpseudocode}
\makeatletter
\renewcommand{\ALG@name}{Pseudocode}
\makeatother
\usepackage{times} 
\usepackage{amsmath} 
\usepackage{amssymb} 

\usepackage{url}
\makeatother

\usepackage{amsthm}

\title{Mathematical model of a remotely controlled skid-slip tracked mobile  robot}

\author{
 Alessia Ferraro, Vito Antonio Nardi \\
 Dipartimento DIIES Ingegneria dell'Informazione e dell'Energia Sostenibile \\
Universit\'a degli studi "Mediterranea" di Reggio Calabria\\
Reggio Calabria\\
  \texttt{alessia.ferraro@unirc.it} \\
   \And
  Valerio Scordamaglia \\
  Dipartimento DIIES Ingegneria dell'Informazione e dell'Energia Sostenibile \\
 Universit\'a degli studi "Mediterranea" di Reggio Calabria\\
  Reggio Calabria\\
}

\begin{document}
\maketitle

\begin{abstract}
In this paper, an uncertain norm-bounded mathematical model for a remotely controlled  skid-slip tracked mobile robot.
The linear state space description aims to describe the nonlinear error dynamics of the robot during the trajectory tracking maneuver in the presence of a delay in the control channel, taking into account  unknown but bounded slip coefficients.
\end{abstract}


\section{Robot mathematical model}\label{SecIItotale}

\begin{figure}
\centering
\epsfig{figure=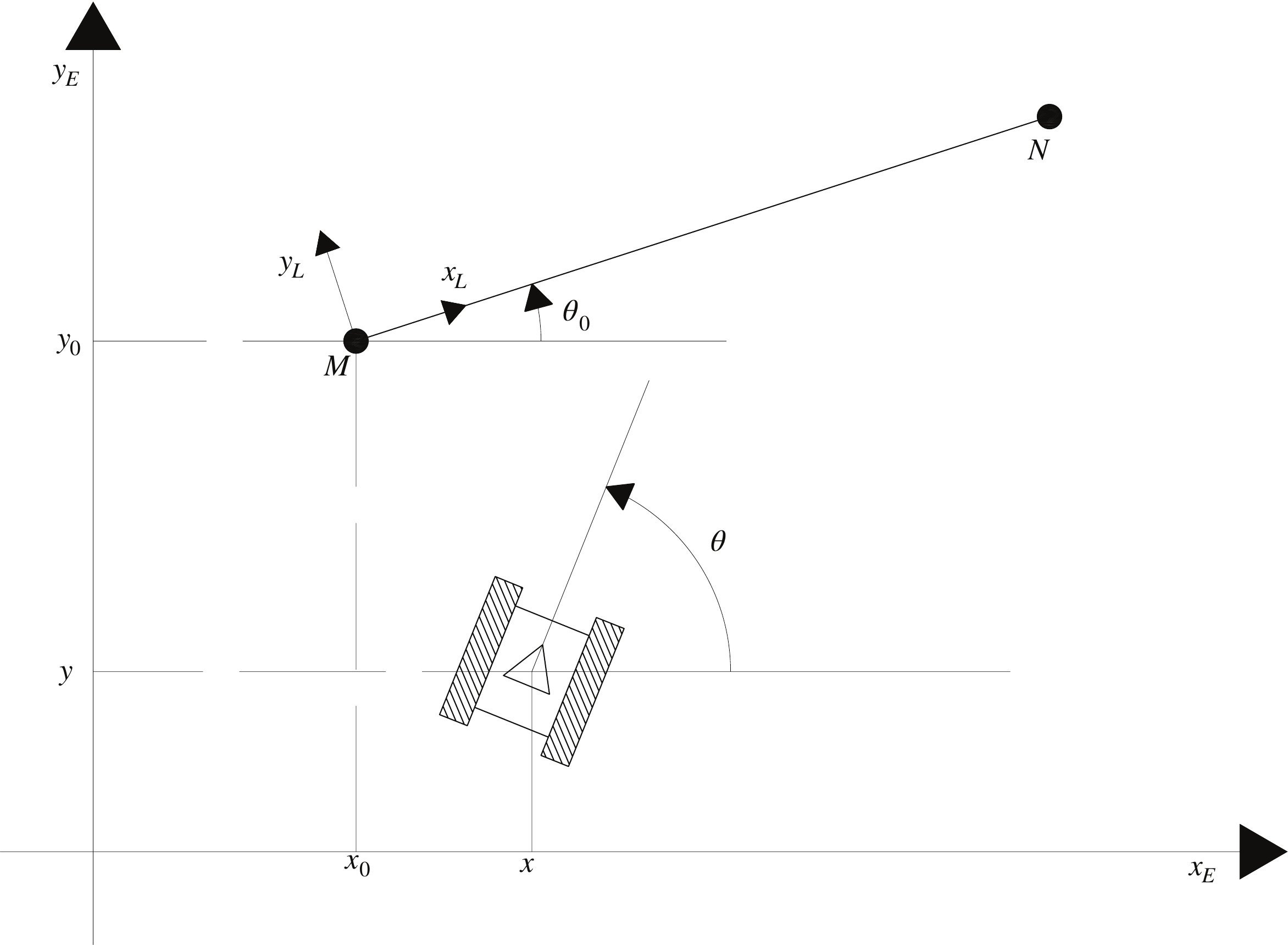,width=\columnwidth}\caption{Reference frames. Robot pose in the $\bf{E}$ frame is denoted as $\{x,y,\theta\}$, the $\bf{L}$ reference frame for trajectory tracking has its origin in $M$ with the $x$-axis oriented in accordance with the assigned trajectory segment $\overline{MN}$}\label{fig.1}
\end{figure}

To define the robot's mathematical model, we assume it can be thought of as a rigid body moving in a horizontal plane.  Given an inertial reference system $\bf{NWU}$ denoted by $\bf{E}$, the robot's pose at time $t$ is 

\begin{equation}\label{eq.1}
q(t)= \begin{bmatrix}
x(t) \,\,\,\,  y(t)  \,\,\,\, \theta(t)
\end{bmatrix}^T
\end{equation}
\noindent where $\theta$ represents the direction and $x$ and $y$ represent the robot's position in $\bf{E}$.
Be
\begin{equation} u(t)=  \begin{bmatrix}
V(t) \,\,\,\ \omega(t)
\end{bmatrix}^T 
\end{equation}
the control velocity vector where the forward and rotational speeds are $V$ and $\omega$ respectively. The following is the first-order kinematic model
\begin{equation}\label{eq.2}
\dot{q}(t) = G(t) \cdot u(t)
\end{equation}
\noindent with
\begin{equation}
G(t) = \begin{bmatrix}
cos(\theta(t)) && 0\\
sin(\theta(t)) && 0 \\
0 && 1 
\end{bmatrix}
\end{equation}
The interaction between tracks and ground may be too complex  to describe mathematically to define a control-oriented mathematical model. Therefore, this interaction has been modeled macroscopically in this paper by using two dimensionless positive time-varying coefficients $\mu_r(t)$ and $\mu_l(t)$ for the right and left tracks,respectively, now referred to as sliding coefficients. The forward and rotational velocities of the robot are related to the $\rho_r(t)$ and $\rho_l(t)$ motor angular velocities through the following relation

\begin{equation} \label{U_t}
u(t) =  J \cdot H(t) \cdot \rho(t)
\end{equation}

being

\begin{equation}
\rho (t)= \begin{bmatrix}
\rho_r(t) \,\,\,\, \rho_l(t)
\end{bmatrix}^T
\end{equation}

\begin{equation}
J= \begin{bmatrix}
R/2 & R/2 \\ 
R/D & -R/D
\end{bmatrix}
\end{equation} 

and

\begin{equation}\label{H_mu}
H(t)=\begin{bmatrix}
\mu_r(t) & 0\\
0 & \mu_l(t)
\end{bmatrix}
\end{equation}
where $R$ is the radius of the gears connecting the tracks to the motors and $D$ is the distance between the tracks.
Let us assume that the robot is controlled by two commands the forward speed $\hat{V}$ and the rotational speed $\hat{\omega}$
\begin{equation} \label{U_hat} 
\hat{u}(t)=\begin{bmatrix} \hat{V}(t) \, \, \hat{\omega}(t) \end{bmatrix}^T
\end{equation} 
Nominal angular speeds of the two electric motors required to generate (\ref{U_hat}) are calculated assuming the two sliding coefficients are both taken as unit values
\begin{equation}\label{eq.9}
\hat{\rho}(t)=   J^{-1}\cdot \hat{u}(t)
\end{equation} 
However, since the sliding coefficients $\mu_r(t)$ and $\mu_l(t)$ depend on the interaction between the track and the ground and may vary over time, the effective velocity of the robot result

\begin{equation}\label{eq.11}
{u}(t)= J\cdot H(t) \cdot \hat{\rho}(t)
\end{equation}
In view of eqs. (\ref{eq.9})-(\ref{eq.11}), eq. (\ref{eq.2}) can be rewritten in the following form
\begin{equation}\label{eq.12}
\dot{q}(t)=G(t)\cdot J \cdot H(t) \cdot J^{-1} \cdot \hat{u}(t)
\end{equation} 
Consider Fig. \ref{fig.1}.
\noindent Let $\bf{L}$ be a reference system centered in $M$ with the coordinates $(x_0,y_0)$ which is aligned with the axis $x$ according to the trajectory segment $\overline{MN}$.
The roto-translation to go from $\bf{E}$ to $\bf{L}$ is
\begin{equation}\label{eq.13}
q_L(t) = R_E^L ( \theta_0) (q(t)-q_0)
\end{equation}
where  $q_0=\begin{bmatrix} x_0 \,\,\, y_0 \,\,\, \theta_0 \end{bmatrix}^T$  with $\theta_0$ represented in Fig. \ref{fig.1} and being
\begin{equation}
R_E^L(\theta_0)= 
\begin{bmatrix}
cos(\theta_0) && sin(\theta_0) && 0 \\
-sin(\theta_0) && cos(\theta_0) && 0 \\
0 && 0 && 1 
\end{bmatrix}
\end{equation}
Be
\begin{equation}
T_L^D(\cdot)=\begin{bmatrix}
q_L{^D(\cdot)}^T && {u^D}^T 
\end{bmatrix}^T
\end{equation}
the desired trajectory expressed in the reference frame $\bf{L}$, defined in terms of permissible pairs of poses and control actions consistent with the kinematic relation (\ref{eq.2}) being $q_L^D(0) =\begin{bmatrix}
0 \,\,\ 0 \,\,\ 0
\end{bmatrix}^T $ and $u^D = \begin{bmatrix} V^D \,\,\ 0 \end{bmatrix}^T$ where $V^D$ is the desired nominal feedforward speed of the robot in the trajectory segment $\overline{MN}$. A null nominal speed is assumed as  rotational speed.
At the time instant  $t$, the desired pose expressed in the reference system $\bf{L}$ is calculated as follows
\begin{equation}\label{eq.16}
q_L^D(t) = \begin{bmatrix}
V^D \cdot t \,\,\ 0 \,\,\ 0 \end{bmatrix}^T
\end{equation}\\
Be 
\begin{equation}\label{eqq.17}
e(t)=q_L(t)-q_L^D(t)= \begin{bmatrix}
e_x(t) \,\,\ e_y(t) \,\,\ e_\theta(t)
\end{bmatrix}^T
\end{equation}
the trajectory tracking error.
Be $\delta u(t)= \hat{u}(t)-u^D=\begin{bmatrix}
\delta V(t) \,\,\ \omega(t) 
\end{bmatrix}^T $, the difference between the effective and nominal control values.
Be  $d(t) = \mu(t)-\bf{1}= \begin{bmatrix}
\delta \mu_r(t) \,\,\ \delta \mu_l(t)
\end{bmatrix}^T$  the deviation of the sliding coefficient values  from the nominal values.
By recombining eqs. (\ref{eq.12}), (\ref{eq.13}) and (\ref{eq.16}), it is possible to rewrite the dynamics of the trajectory tracking error 
\begin{equation}\label{eq.17}
\dot{e}(t) =  R_E^L(\theta_0) \cdot G(t) \cdot J \cdot H(t) \cdot J^{-1} \cdot \hat{u}(t) - \begin{bmatrix}
V^D \\
0 \\
0
\end{bmatrix}
\end{equation}
Finally, by applying the classical linearization procedure, it is possible to define the following linear time-invariant representation around the nominal condition $\Sigma(t)=\{q_L^D(t), \,\,\,u^D\}$

\begin{equation}\label{lineare}
\dot{e}(t) = A e(t) +B \delta u(t) +B_{D} d(t)
\end{equation}
\begin{equation}
 A=\frac{\partial \dot{e}(t)}{\partial{q}_L(t)}\bigg|_{\Sigma(t)}=
\begin{bmatrix}
0&0&0\\
0&0&V^D\\
0&0&0\\
 \end{bmatrix}
\end{equation}
\begin{equation}
 B=\frac{\partial \dot{e}(t)}{\partial{\hat{u}(t)}}\bigg|_{\Sigma(t)}=\begin{bmatrix}
1&0\\
0&0\\
0&1
 \end{bmatrix}
\end{equation}
\begin{equation}
B_{D}=\frac{\partial \dot{e}(t)}{\partial{\mu(t)}}\bigg|_{\Sigma(t)}=\begin{bmatrix}
\frac{V^D}{2}&\frac{V^D}{2}\\
0&0\\
\frac{V^D}{D}&-\frac{V^D}{D}
\end{bmatrix}
\end{equation}

\subsection{Modelling of networked control system}\label{SecIIbis}
\begin{figure}
\centering
\epsfig{figure=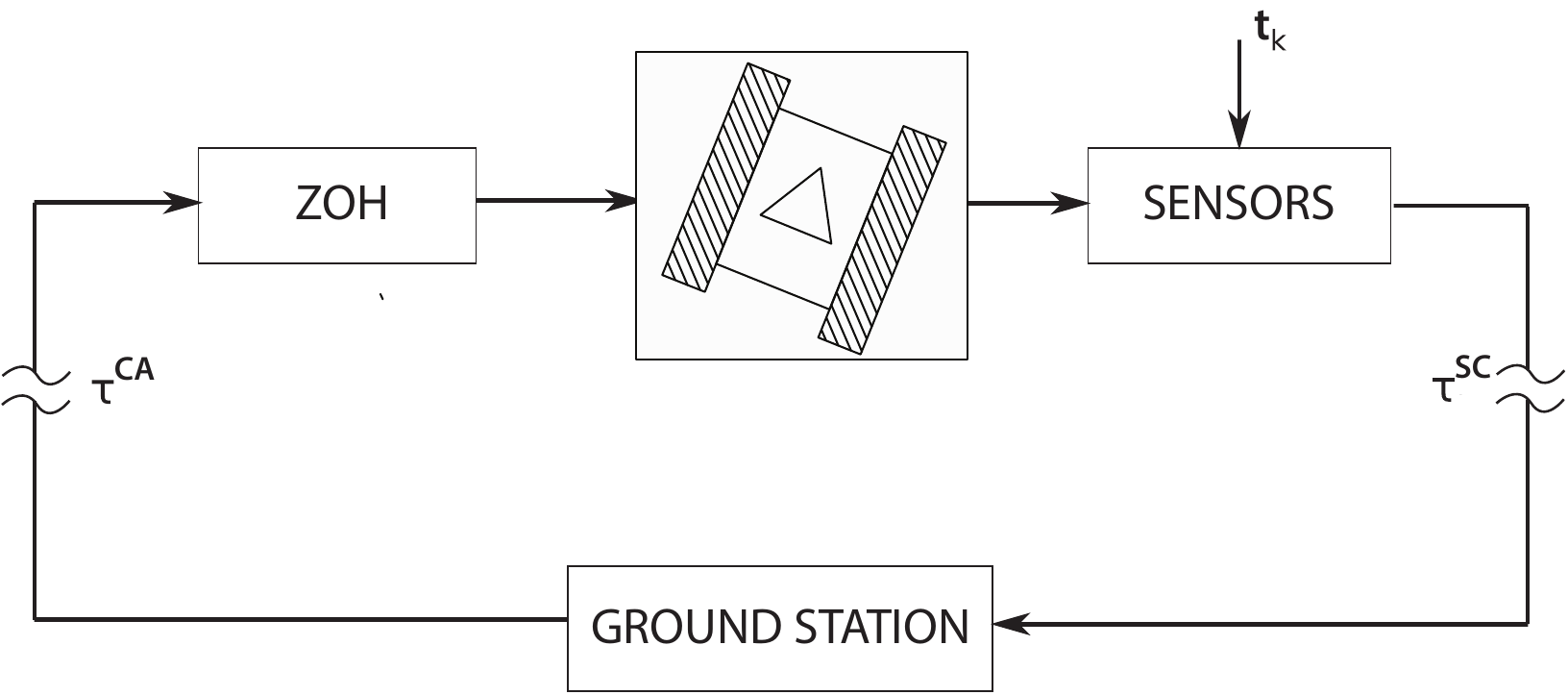,width=\columnwidth}\caption{An outline of a NCS control schema.\label{schemaNCS}}
\end{figure}

Assume that the robot is controlled remotely via a data communication network. The generic control scheme is shown in Fig. \ref{schemaNCS}. This solution offers numerous advantages. Most important is the simplicity of installation and maintenance of the control system as well as the high flexibility. However, despite the positive aspects, the presence of a data communication network on a wireless or wired channel within a control loop can lead to a degradation of system performance (up to instability) due to phenomena such as packet loss and time delays. With the theory of Networked Control Systems \cite{bemporad2010networked}, it is possible to mathematically model the presence of a data communication channel in the control loop.

Suppose you are using a connection-oriented communication protocol (e.g., TCP-IP) where the number of packets lost during communication is assumed to be zero. Let $\tau^{SC}$ be the \emph{sensors-to-control} delay, which represents the time it takes to transmit information from the sensor to the controller. Let $\tau^{CA}$ be the \emph{control-to-actuator} delay, which represents the time it takes for the control signal to travel from the controller to the actuator. Finally, consider an additional delay $\tau^C$, which is the computation time it takes for the controller to process the new control strategy. In this work, all other delays in the control loop are considered negligible.

\begin{figure}
\centering
\epsfig{figure=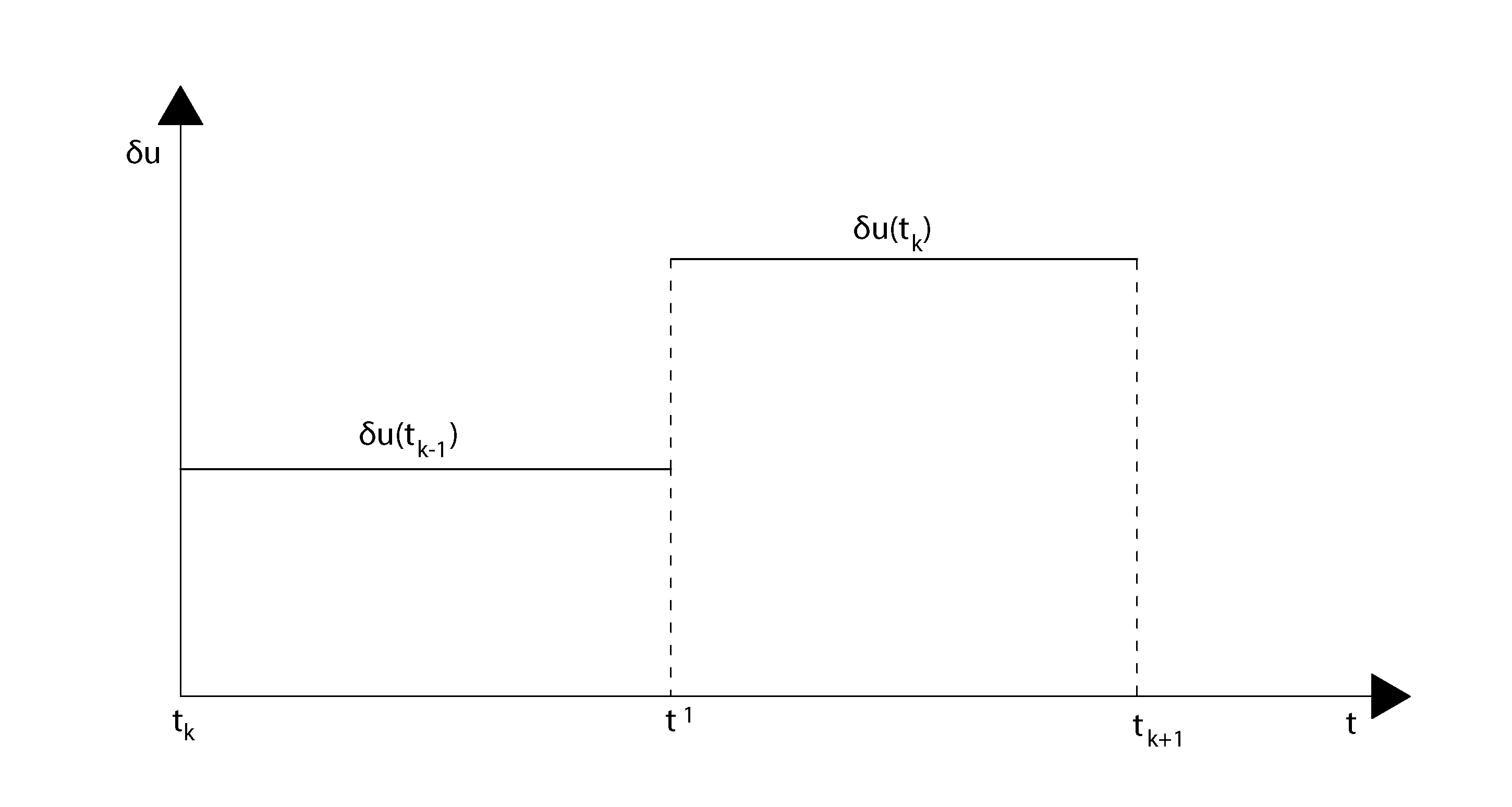,width=\columnwidth}\caption{Temporal diagram of actuation update}\label{Ncs_2}
\end{figure}

Let us assume that the sensors operate synchronously: sampling occurs at times $t_k= k\cdot T_s$ with $k \in N$ and $T_s$ is the sampling time. Let us assume that both the controller and the actuators are event-driven, i.e., they react immediately to new data. Under these assumptions, the three delays mentioned above can be conveniently combined into a single delay term

\begin{equation}\label{network_delay}
\tau=\tau^{SC}+\tau^{CA}+\tau^{C}
\end{equation} 
which is assumed may change over time. 
Let $\tau_{max}$ ($\tau_{min}$) be the maximum (minimum) possible value of $\tau$. Let $\bar{d}$ be the smallest positive integer satisfying the relation $\bar{d}\ge\tau_{max}/T_s$. Let $\underline{d}$ be defined as the largest positive integer satisfying the relation $\underline{d}\le \tau_{min}/T_s$. During a generic sampling period $T_s$, the control action can change at most $\bar{d}-\underline{d}$ times \cite{heemels2010comparison}. For simplicity, we will assume that $0 \leq \tau_{min} \leq \tau_{max} \leq T_s$ such that $\bar{d}=1$ and $\underline{d}=0$. So consider Fig. \ref{Ncs_2} with $T_s-\tau_{max} \le t_{k+1}-t^1 \le T_s-\tau_{min}$.

The state of the system (\ref{lineare}) at time $t_{k+1}$ can be rewritten as 

\begin{equation} \label{NCS_tk+1}
e(t_{k+1})={\rm{e}}^{A(t_{k+1}-t^1)}e(t^1)+\int_0^{t_{k+1}-t^1}{\rm{e}}^{A \sigma}d\sigma B\delta u(t_k)+\int_0^{t_{k+1}-t^1}{\rm{e}}^{A \sigma}d\sigma B_{D}d(t_k)
\end{equation}

In the same manner 

\begin{equation} \label{NCS_tauk}
e(t^1)={\rm{e}}^{A(t^1-t_k)}e(t_k)+\int_0^{t^1-t_k}{\rm{e}}^{A \sigma}d\sigma B\delta u(t_{k-1})+\int_0^{t^1-t_k}{\rm{e}}^{A \sigma}d\sigma B_{D}d(t_k)
\end{equation}

By appropriately recombining (\ref{NCS_tk+1}) and (\ref{NCS_tauk}), the following form can be obtained 

\begin{align}\label{LTI_error2}
&e(t_{k+1})={\rm{e}}^{A T_s} e(t_k) + \int_0^{t_{k+1}-t^1}{\rm{e}}^{A \sigma}d\sigma B\delta u(t_k)+ \int^{Ts}_{t_{k+1}-t^1}{\rm{e}}^{A \sigma}d\sigma B\delta u(t_{k-1}) + \\ \nonumber
&+\int_0^{Ts} {\rm{e}}^{A \sigma}d\sigma B_{D} d(t_k)
\end{align}
For sake of clarity, the time-dependent functions sampled at time instants $t _k$ will henceforth be denoted by the notation $f_k=f(t_k)$.

Be
\begin{equation}\label{liftedstate}
\tilde{\xi}_k = \begin{bmatrix}e_k^T & \delta u_{k-1}^T \end{bmatrix}^T\in\mathcal{R}^{n_s}
\end{equation} 
the vector of lifted states, hereinafter named \emph{lifted trajectory tracking error}, being $n_s=n_e+(\bar{d}-\underline{d})\cdot n_u$ with $n_e$ and $n_u$ the number of states and inputs of (\ref{lineare}) respectively (in this case $n_e=3$ and $n_u=2$).
Eq. (\ref{LTI_error2}) can be rewritten in the following form

\small  
\begin{equation}\label{eq.25}
\tilde{\xi}_{k+1} =\underbrace{ \begin{bmatrix}
{\rm{e}}^{A T_s}  && \int^{Ts}_{t_{k+1}-t^1}{\rm{e}}^{A \sigma}d\sigma B \\
\bf{0} && \bf{0} 
\end{bmatrix}}_{\tilde{A}(t_{k+1}-t^1)} \tilde{\xi}_k + \underbrace{\begin{bmatrix}
\int_0^{t_{k+1}-t^1}{\rm{e}}^{A \sigma}d\sigma B \\
I
\end{bmatrix}}_{\tilde{B}(t_{k+1}-t^1)} \delta u_(t_k) +\underbrace{\begin{bmatrix}
\int_0^{Ts} e^{A \sigma}d\sigma B_{D} \\
\bf{0}
\end{bmatrix}}_{\tilde{B}d} d_k
\end{equation} 
\normalsize
Finally, system (\ref{eq.25}) can be embedded  in the following norm-bounded uncertain representation  

\begin{equation}\label{eq31}
\tilde{\xi}_{k+1}=\tilde{A}\tilde{\xi}_k+\tilde{B}\delta u_k+\tilde{B}_D d_k+\tilde{B}_pp_k
\end{equation}
\begin{equation}\label{eq32}
p_k=\Delta_k q_k
\end{equation}
\begin{equation}\label{eq33}
q_k=\tilde{C}_q\tilde{\xi}_{k}+\tilde{D}_q  \delta u_k
\end{equation}
where $\|\Delta_k \|< 1\,\, \forall k\ge 0$, and $\tilde{B}_p$, $\tilde{C}_q$ and $\tilde{D}_q$ matrices of proper dimensions.
It is worth noting that the same approach can be used to define an uncertain representation with norm-bounded uncertainty for different maximum and minimum value of the delay $\tau$.

\bibliographystyle{unsrt}  
\bibliography{fdi_biblio}

\end{document}